\documentclass[10pt,twocolumn,letterpaper]{article}

\usepackage{cvpr}
\usepackage{times}
\usepackage{epsfig}
\usepackage{graphicx}
\usepackage{amsmath,amsfonts,amsthm,amssymb}
\usepackage{setspace}
\usepackage{fancyhdr}
\usepackage{lastpage}
\usepackage{extramarks}
\usepackage{chngpage}
\usepackage{soul,color}
\usepackage{graphicx,float,wrapfig}
\usepackage[boxed, linesnumbered]{algorithm2e}
\usepackage{algorithmicx}
\usepackage{dsfont}
\usepackage{subfigure}
\usepackage{listings}
\usepackage{enumerate}
\usepackage{multirow}
\usepackage{framed}
\usepackage{enumitem}

\renewcommand{\Re}{\mathbb{R}}

\usepackage[pagebackref=true,breaklinks=true,letterpaper=true,colorlinks,bookmarks=false]{hyperref}

\cvprfinalcopy 

\def\cvprPaperID{1512} 

\ifcvprfinal\fi
\begin{document}

\title{SparkleVision: Seeing the World through Random Specular Microfacets}

\author{
Zhengdong Zhang, Phillip Isola, and Edward H. Adelson\\
Massachusetts Institute of Technology\\
{\tt\small \{zhangzd, phillipi, eadelson\}@mit.edu}
}

\maketitle

\begin{abstract}
In this paper, we study the problem of reproducing the world lighting from a single image of an object covered with random specular microfacets on the surface. We show that such reflectors can be interpreted as a randomized mapping from the lighting to the image. Such specular objects have very different optical properties from both diffuse surfaces and smooth specular objects like metals, so we design special imaging system to robustly and effectively photograph them. We present simple yet reliable algorithms to calibrate the proposed system and do the inference. We conduct experiments to verify the correctness of our model assumptions and prove the effectiveness of our pipeline.
\end{abstract}

\section{Introduction}
An objects appearance depends on the properties of the object itself as well as the surrounding light. How much can we tell about the light from looking at the object? If the object is smooth and matte, then we can tell rather little \cite{LambertianBJ,LambertianRH,LambertianRH2,LambertianRH3}. However, if the object is irregular and/or non-matte, there are more possibilities.

\begin{figure}[!t]
\centering{
    \subfigure[Specular microfacets]{
        \includegraphics[height=0.23\linewidth]{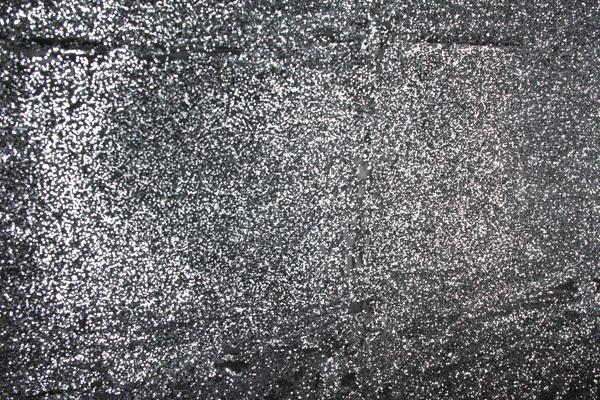}
    }
    \subfigure[Closer look]{
        \includegraphics[height=0.23\linewidth]{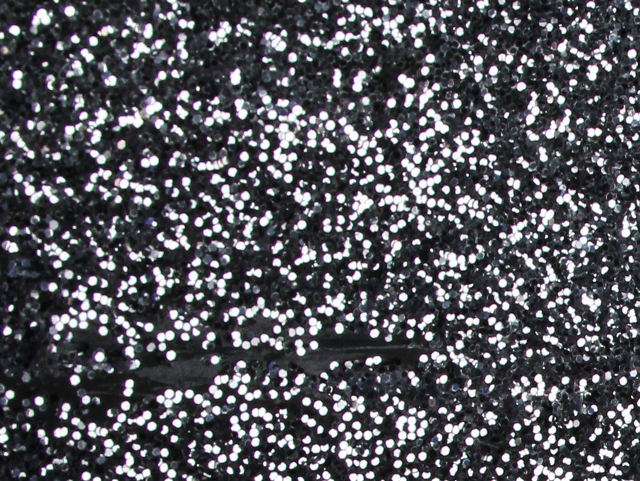}
    }
    \subfigure[Reconstructed lighting]{
        \includegraphics[height=0.23\linewidth]{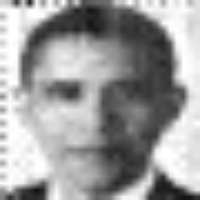}
    }
}
\caption{Reproducing the world from a single image of specular random facets: (a) shows the image of a surface covered with glitter illuminated by a screen showing the image of Obama. (b) gives a close up look of (a), highlighting both the bright spots and dark spots. (c) shows the lighting, i.e., the face of Obama the our algorithm constructs from (a).}
\label{fig:specular_teaser}
\end{figure}

Figure \ref{fig:specular_teaser} shows a picture of a surface covered in glitter. The glitter is sparkly, and the image shows a scattering of bright specularities. We may think of the glitter as containing mirror facets randomly oriented. Each facet reflects light at a certain angle. If we knew the optical and geometrical properties of the facets, we could potentially decode the reflected scene.

\begin{figure*}[!tb]
\centering{
    \subfigure[flat mirror]{
    	\label{fig:flat_mirror}
        \includegraphics[width=0.18\linewidth]{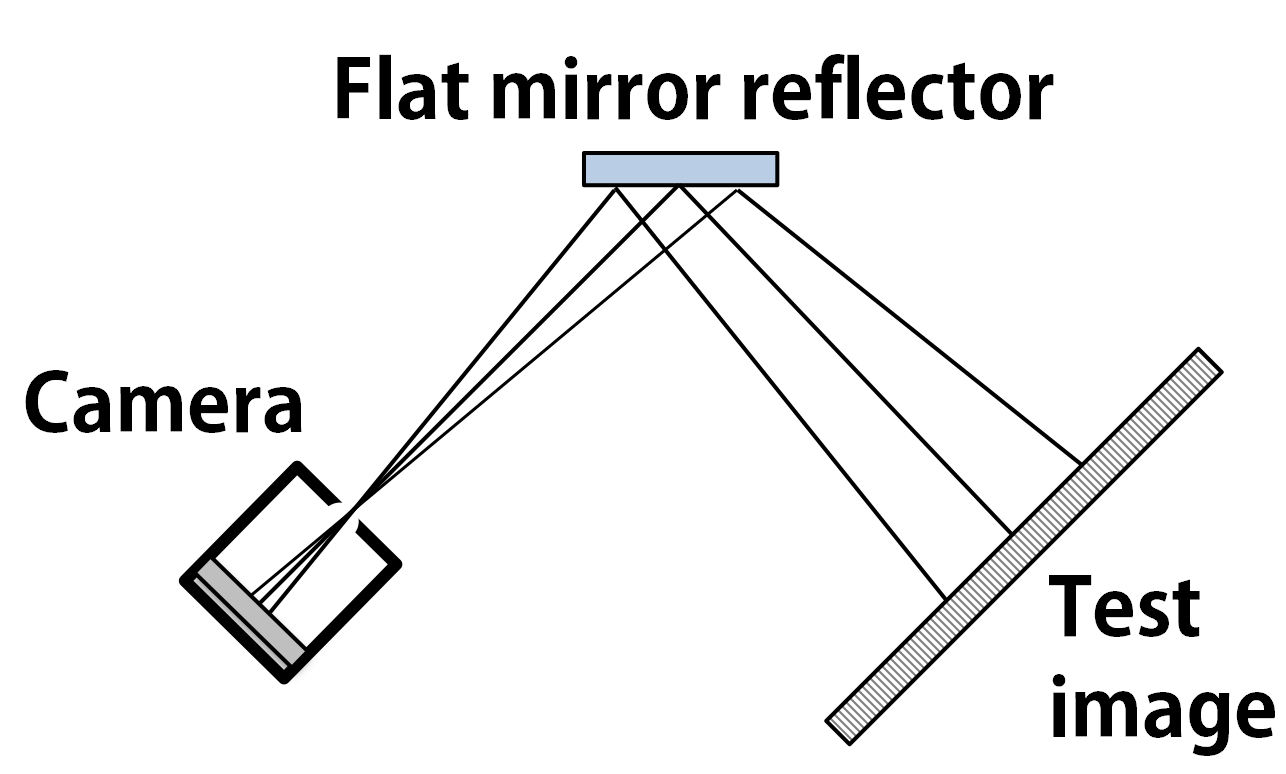}
    }
    \subfigure[curved mirror]{
    	\label{fig:curved_mirror}
        \includegraphics[width=0.18\linewidth]{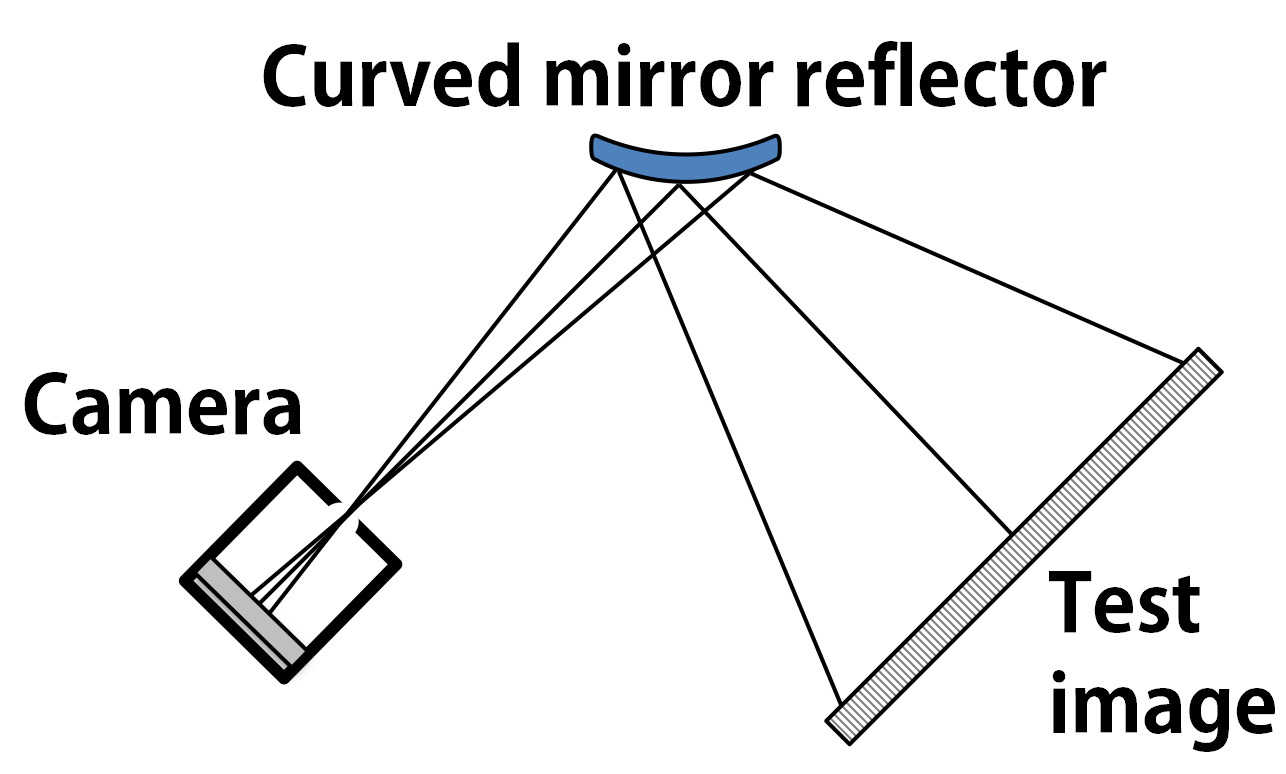}
    }
    \subfigure[smashed mirror, \emph{sparkle-vision}]{
    	\label{fig:smashed_mirror}
        \includegraphics[width=0.18\linewidth]{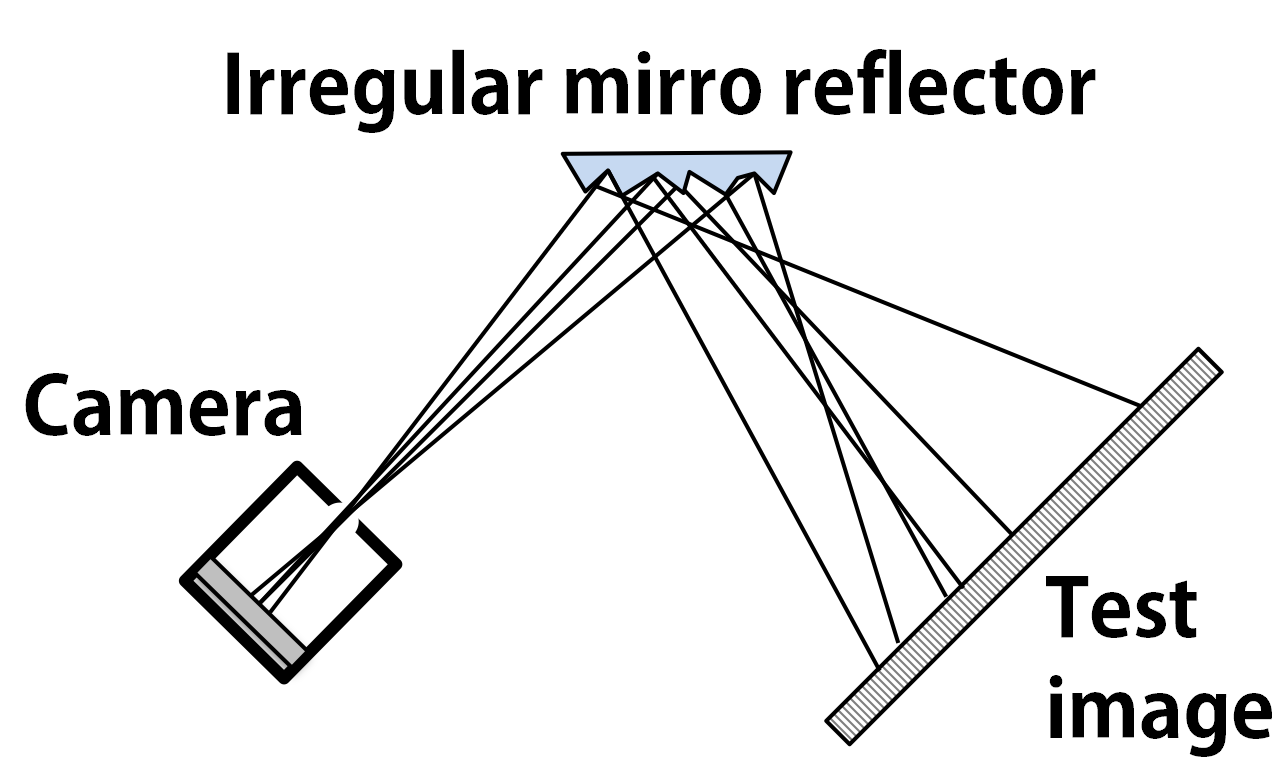}
    }
    \subfigure[irregular matte reflector]{
    	\label{fig:irregular_matte_reflector}
        \includegraphics[width=0.18\linewidth]{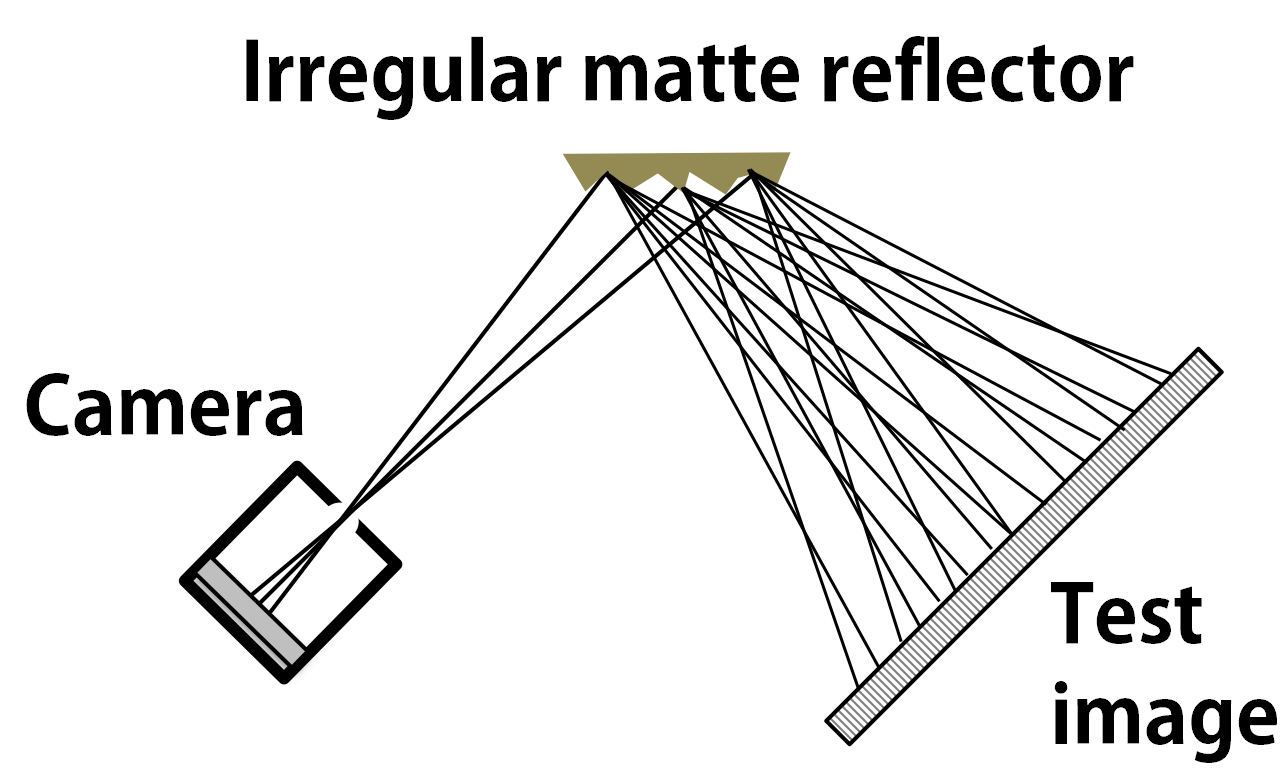}
    }
    \subfigure[irregular mirror without pinhole camera]{
    	\label{fig:irregular_mirror_no_pinhole}
        \includegraphics[width=0.18\linewidth]{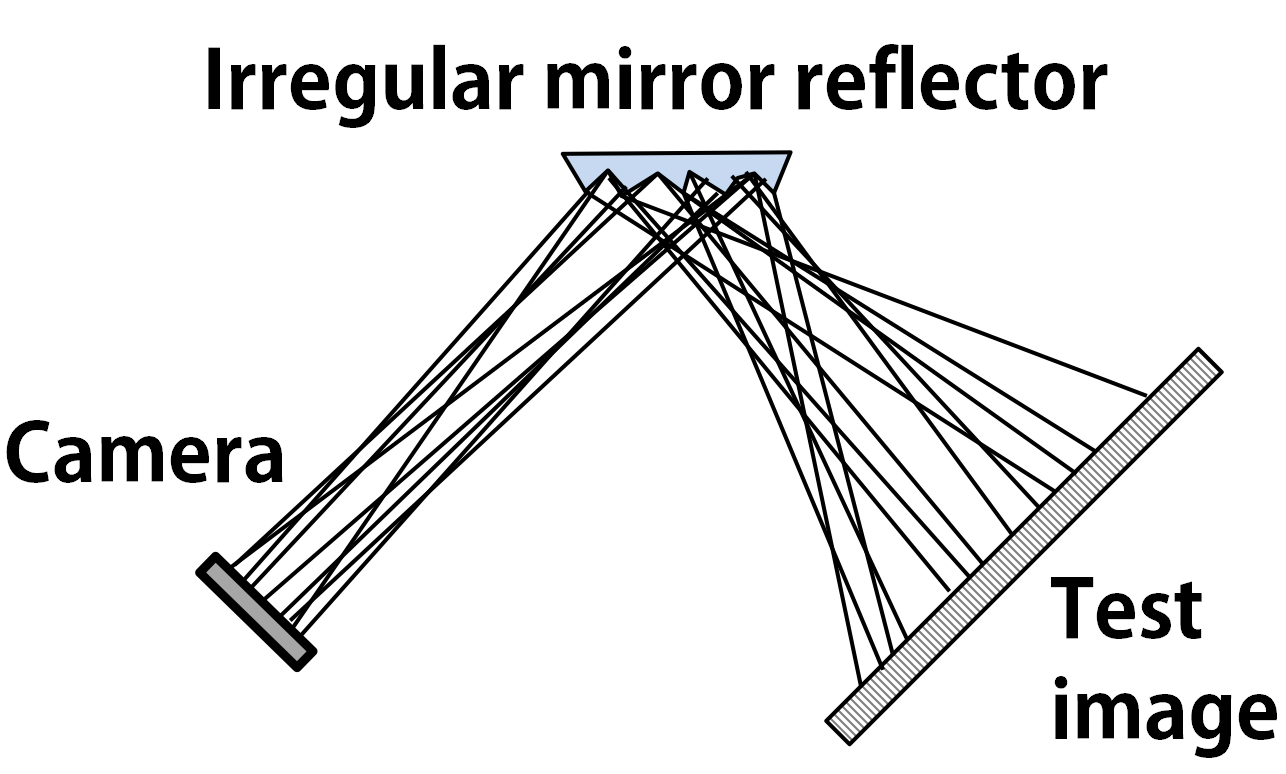}
    }
}
\caption{Optical arrangements in which light rays travel from a scene to a camera sensor by way of a reflector: ``SparkleVision'' refers to the setup in (c).}
\label{fig:optical_systems}
\end{figure*}

Figure \ref{fig:optical_systems} shows a variety of optical arrangements in which light rays travel from a scene to a camera sensor by way of a reflector. For simplicity we assume the scene is planar; for example it could be a computer display screen showing a test image. A subset of rays are seen by the sensor in the camera. Here we show a pinhole camera for simplicity.

Figure \ref{fig:flat_mirror} shows the case of an ordinary flat mirror reflector. The pinhole camera forms an image of the display screen (reflected in the mirror) in the ordinary way. There is a simple mapping between screen pixels and sensor pixels. Figure \ref{fig:curved_mirror} shows the same arrangement with a curved mirror. Again there is a simple mapping between screen pixels and sensor pixels. The field of view is wider due to the mirror's curvature. Figure \ref{fig:smashed_mirror} shows the case of a smashed mirror, which forms an irregular array of mirror facets. The ray directions are scrambled, but the mapping between screen pixels and sensor pixels is still relatively simple. This is the situation we consider in the present work.

Figure \ref{fig:irregular_matte_reflector} shows the case of an irregular matte reflector. Each sensor pixel sees a particular point on the matte reflector, but that point integrates light from a broad area of the display screen. Unscrambling the resulting image is almost impossible, although there are cases where some information may be retrieved, as shown by \cite{AccidentalPinholeCamera} in their discussion of accidental pinhole cameras. Figure \ref{fig:irregular_mirror_no_pinhole} shows the case of an irregular mirror, but without benefit of a pinhole camera restricting the rays hitting the sensor. This case corresponds to the random camera proposed by Fergus et al \cite{RandomLens}, in which the reflector itself is the only imaging element. Since each pixel captures light from many directions, unscrambling is extremely difficult.

The case in Figure \ref{fig:smashed_mirror}, with a sparkly surface and a pinhole camera, deserves study. We call this case ``SparkleVision". It involves relatively little mixing of light rays, so unscrambling seems feasible. Moreover it could be of practical value, since irregular specular surfaces occur in the real world (e.g., with metals, certain fabrics, micaceous minerals, and the Fresnel reflections from foliage or wet surfaces).

For a surface covered in glitter, it is difficult to build a proper physical model. Instead of an explicit model, we can describe the sparkly surface plus camera as providing a linear transform on the test image. With a planar display screen, each sparkle provides information about some limited parts of the screen. Non-planar facets and limited optical resolution will lead to some mixture of light from multiple locations. However, the transform is still linear. There exists a forward scrambling matrix, and in principle we can find its inverse and unscramble the image.

To learn the forward matrix we can probe the system by displaying a series of test images. These could be orthogonal bases, such as a set of impulses, or the DCT basis functions. They could also be non-orthogonal sets, and can be overcomplete.  Having determined the forward matrix we can compute its inverse.

All the optical systems shown in Figure \ref{fig:optical_systems} implement linear transforms, and all can be characterized in the same manner. However, if a system is ill-conditioned, the inversion will be noisy and unreliable. The performance in practice is an empirical question. We will show that the case in Figure \ref{fig:smashed_mirror}, SparkleVision, allows one to retrieve an image that is good enough to recognize objects and faces.

\subsection{Related Work}

A diffuse object like a ping pong ball tells us little about the lighting. If a Lambertian object is convex, its appearance approximately lies in a nine-dimensional subspace \cite{LambertianBJ,LambertianRH,LambertianRH2,LambertianRH3}, making it impossible to reconstruct more than a $3\times3$ environment map. For non-convex objects, an image of the object under all possible lighting conditions lies in a much higher dimension space due to shadows and occlusions\cite{JohnGuaranteed}, enabling the reconstruction of light beyond $3\times 3$ \cite{DiffuseCamera}. But in general, matte surfaces are tough to work with.

A smooth specular object like a curved mirror provides a distorted image of the lighting, which humans can recognize \cite{Blake1990SpecularBrain} and algorithms can process \cite{SpecularTheory,SpecularLinear,SpecularFlow}. However, specular random facets are different. Typically they are highly irregular and discontinuous, making it hard even for humans to perceive. We utilize a similar model with inverse light transport \cite{LightTransportSMK,LightTransport2} to analyze this new setup, and propose a novel pipeline to effectively reduce the noise and increase the stability of the system, both in calibration and reconstruction.


Researchers have applied micro-lens arrays to capture lightfields \cite{adelson1992single, levoy2006light}. To some extent, a specular reflector can also be considered as a coded aperture of a general camera system\cite{levin2007image}. Our work differs from the previous work in the sense that our setup is randomized -- to the best of our knowledge previous work in this domain mainly uses specially manufactured array with known mapping whereas in our system the array is randomly distributed.

Many ideas in this paper are inspired by previous work on Random Camera \cite{RandomLens}. However, the key difference between our paper and the previous work is that in \cite{RandomLens} no lens is used and hence all the lights from all directions in the lightfield get mixed up which is difficult to invert. In our setup, we place a lens between the world and the camera sensor, which makes the problem significantly easier and more tractable to solve. Also similar ideas appears in the ``single pixel camera'' \cite{SinglePixel} where measurements of the light are randomized for compressed sensing.

The idea that some everyday objects can accidentally serve as a camera has been explored before. It is in shown in \cite{EyeRelighting} that an photograph of a human eye reflects the environment in front of the eye, and this can be used for relighting. In addition, a window or a door can act like a pinhole, in effect imaging the world outside the opening\cite{AccidentalPinholeCamera}.


\section{The formulation of SparkleVision}
\label{sec:notation}
We discuss the optical setup of SparkleVision in discretized settings.  Suppose the lightfield in a particular environment is denoted by a stacked, discrete vector $x$ in the space. We place a specular object $O$ with random specular microfacets into the environment. Further we use a camera $C$ with a focused lens to capture the intensity of the light reflected by $O$. Let the discrete vector $y$ be the sensor output. It is well known that any passive optical system is linear.  So we use a matrix $\mathcal{A}(\cdot)$ to represent the linear mapping relating the lightfield $x$ to $y$. Therefore, $y = Ax$.

Note that all the above discussion makes no assumption on any material, albedo, smoothness or continuity properties of the objects in the scene. Therefore, this linear representation holds for any random specular microfacets. In this notation, the task of \emph{SparkleVision} can be summarized as
\begin{itemize}[noitemsep,nolistsep]
\item Accurately capture the image $y$ of a sparkling object.
\item Determine the matrix $A$, which is a calibration task.
\item Infer the light $x$ from the image $y$, which is an inference task.
\end{itemize}

In the later discussion, we will use many pairs of lightings and images so we use the subscript $(x_i, y_i)$ to denote the $i$-th pair of them. In addition, let $e_i$ be the $i$-th unit vector of the identity basis, i.e., a vector whose entries are all zero except the $i$-th entry which is one. Similarly, let $d_i$ represent the $i$-th unit vector of the bases of the Discrete Cosine Transform (DCT). We use $b_i$ to represent a random basis vector where all entries are i.i.d random variables. Also let $A = [a_1, a_2, \ldots, a_N]$ with $a_i$ as its $i$-th column.

\section{Imaging specular random facets through HDR}
\label{sec:imaging}
In this section we examine the properties of sparkling objects with microfacets. Their special characteristics enable the recovery of lighting from an image while imposing unique challenges to accurately capture images of them. To deal with these challenges, we use High Dynamic Ranging (HDR) imaging, using multiple exposures of the same scene.

\subsection{Sparkling Pattern under Impulse Lightings}

\begin{figure}[!tb]
\centering{
    \subfigure[image]{
    	\label{fig:specular_img_demo}
        \includegraphics[height=0.25\linewidth]{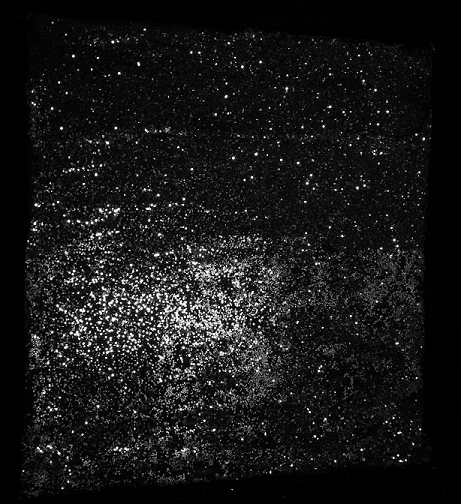}
    }
    \subfigure[histogram]{
    	\label{fig:specular_img_hist}
        \includegraphics[height=0.25\linewidth]{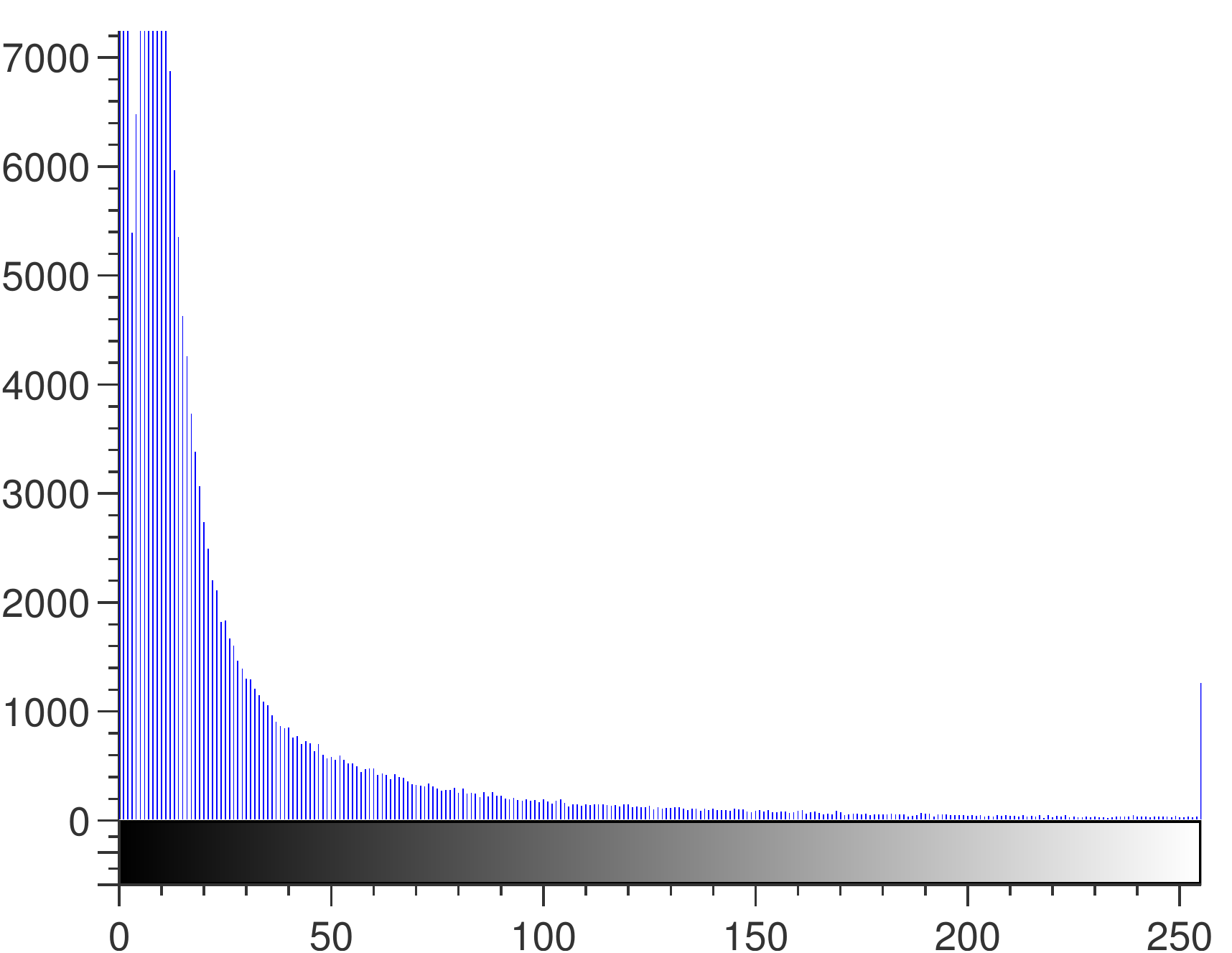}
    }
    \subfigure[non-overlap]{
    	\label{fig:overlap}
        \includegraphics[height=0.25\linewidth]{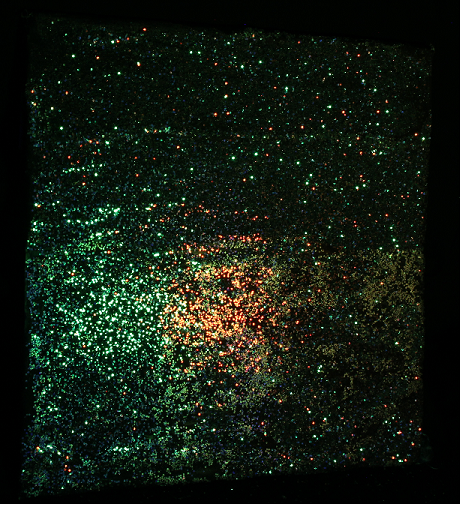}
    }
}
\caption{Optical properties of specular random micro facets: (a) shows an image of specular facet with scattered bright spots. (b) demonstrates its histogram, although the bright spots in the image are shining, most of pixels are actually dark. (c) the surface simultaneously illuminated by two adjacent impulse lighting, one in red and one in green. The reflected green lights and red lights seldom overlap as few spots in the image are yellow.}
\label{fig:specular_property}
\end{figure}

Specular random microfacets can be considered as a randomized mapping between the world light and the camera. Each single facet faces a random orientation. It acts as a mirror reflecting all the incoming lights. However, because of the existence of a camera with focused lens and the small size of each facet, only lights from a very narrow range of directions will be reflected into camera from any given facet. Therefore, given a single point light source, only a very small number of the facets will reflect light to the camera and appear bright. The rest of the facets will be unilluminated. This effect makes the dynamic range of a photo of specular facets extremely high, creating a unique challenge to photographing them. Figure \ref{fig:specular_img_demo} and \ref{fig:specular_img_hist} plots the histogram of a photo of such reflector.

Now suppose we slightly change the location of the impulse light, generating a small disturbance to the direction of the incoming light to the facets. Thanks to the narrow range of reflecting direction of each facet, this slight disturbance will cause a huge change of light patterns on the random facets. Provided that the orientations of the facets are random across all the surfaces, we should expect that the set of aligned facets will be significantly different. Figure \ref{fig:overlap} gives us an illustration of this phenomenon. Intuitively, if our task is just to decide whether a certain point light source is on or not, we could just count whether the corresponding set of facets for that light's position is active or not. This also suggests that our system is very sensitive to the alignment of the geometric setup, which we will address in Section \ref{sec:misalignment}.

\subsection{HDR Imaging}
As we have seen, the dynamic range of an image of a sparkling object is extremely high. Dark regions are noisy and numerous throughout the image. To accurately capture them, long exposure is needed. Unfortunately, long exposure makes the bright spots saturated and therefore breaks the linearity assumption. If we adjust the exposure for the sparse bright spots, the exposure time would be too short to capture the noisy dark pixels accurately. Therefore, it is not practical to capture both high and low intensity illumination with just a single shot with a commercial DSLR camera.

Our solution is to take multiple shots of the same scene with $K$ different exposure time $t_k$. Let the resulting images be $I_1, I_2, \ldots, I_k$. We can then combine those $K$ images into a single image $I_0$ with much higher dynamic range than any of the original K images. Note that we use a heavy tripod in the experiment and therefore we assume all $I_i$ are already registered. Therefore, we only need to develop a way to decide $I_0(x)$ from $I_k(x)$ for any arbitrary location $x$.

The Canon Rebel T2i camera that we use in our experiments has roughly linear response with respect to the exposure time for a fairly large range -- roughly when the intensity ranges in $(0.1, 0.7)$. When the intensity goes beyond $0.7$ the response function becomes curved and gradually saturated and hence the linearity assumption breaks down. When the intensity is lower than $0.1$ the image is very noisy. So we need to discard these undesired intensities. Denote the remaining exposure time and intensity pairs $(t_i, I_i(x))$. The goal is to determine the value $I(x)$ independently for each location $x$. We solve this problem by fitting a least squares line to $(t_i, I_i(x))$:

\begin{equation}
I(r) = \mathrm{argmin}_{s} \sum_i (s\cdot t_i - I_i(x))^2
\end{equation}
With simple algebra we can derive a closed form solution:
\begin{equation}
I(r) = \frac{\sum_i t_i I_i(r)}{\sum_i t_i^2}
\end{equation}
Note that the derived solution can be viewed as an average of intensities under different exposures weighted by the exposure time.

\section{Calibration and Inference of SparkleVision System}
In this section, we examine the algorithm to calibrate the system and reconstruct the environmental map $x$ from $y$.

\subsection{Calibration with overcomplete basis}
We probe the system $y = Ax$ by illuminating the object with impulse lights $e_i$. Ideally, $y_i = A\cdot e_i = a_i$. So we can scan through all $e_i$ to get $A$. However, due to the presence of noise the measured $y_i$ will typically differ from the $a_i$ of an ideal system. As we will show later in experimental sections, this noise on the calibrated matrix $A$ is lethal to the recovery of the light. Our system relies on a clean, accurate transformation matrix $A$ to succeed. Therefore, we further probe the system with multiple different basis. Specifically we use the DCT basis $d_i$ and a set of random basis $b_i$. Doing this we make the system over-complete and hence the estimated $A$ becomes more robust to noise. Let $E$ be the $N\times N$ impulse basis matrix, $D$ be the DCT basis matrix and $B_K\in\Re^{N\times K}$ be the matrix of $K$ random basis vectors. This implies the following optimization to do the calibration:
\begin{small}
\begin{equation}
\label{eqn:naive_calib}
\min_{A} \|Y_1 - AE\|_F^2 + \lambda \|Y_2 - AD\|_F^2 + \lambda \|Y_3 - A B_K\|_F^2
\end{equation}
\end{small}
$\lambda$ here is a weight to balance the error since the illumination from impulse lights tend to be much dimmer than the illumination from DCT and random lighting. In our experiments we set $\lambda = \frac{1}{N}$.

To further refine the quality of the calibrated $A$ against the noise in the dominant dark regions of $A$, we only retain intensities above a certain intensity during calibration. Specifically let $\Omega_i$ be the set of the 1\% brightest pixels in $y_i$ illuminated by the impulse $e_i$. Let $\Omega = \bigcup_i \Omega_i$. We then only keep the pixels inside $\Omega$ and discard the rest. Let $P_\Omega(\cdot)$ represents such a projection. This turns the calibration into the following optimization:
\begin{small}
\begin{equation}
\label{eqn:proj_calib}
\min_{A} \|P_\Omega(Y_1) - AE\|_F^2 + \lambda \|P_\Omega(Y_2) - AD\|_F^2 + \lambda \|P_\Omega(Y_3) - A B_K\|_F^2
\end{equation}
\end{small}

Note that the size of the output $A$ from \eqref{eqn:proj_calib} is different from \eqref{eqn:naive_calib} due to the projection $\Omega(\cdot)$.

\subsection{Reconstruction}
Given $A$, reconstructing an environment map from an image $y$ is a classic inverse problem. A straightforward approach to this problem is to solve it by least-squares. However, this unconstrained least square may produce entries less than 0, which is not physically meaningful. Instead we solve a constrained least squares problem:

\begin{equation}
\label{eqn:recon}
\min_{x} \|y - Ax\|_F^2, \ \ s.t.\ \ x\geq0
\end{equation}

Nevertheless, through experiments we find they are actually too slow for application. When the resolution of the screen is $20\times 20$, i.e., $x\in \Re^{400}$, solving the inequality constrained least square is approximately 100 times slower. Yet the improvement is minor. So we just solve the naive least square without non-negative constraints and then crop the out-ranged pixels back to $[0, 1]$.

We observe that in practice there is room for improvement to smooth the outcome of the above optimization. For example, we could impose stronger image priors to make the result more visually appealing. However, doing so would disguise some of the intrinsic physical behavior of SparkleVision, and therefore we decide to stick to the most naive optimization \eqref{eqn:recon}.

\subsection{Extensions and implementation details}
We use RAW files from the camera to avoid any non-linearity post-processing in image format like JPEG and PNG. In addition, we model the background light as $e_0$ and shot $y_0 = Ae_0$ by turning all active light sources off. We subtract $y_0$ from every $y_i$ in the experiments by default. Since $y_0$ is used for all of $y_i$, we repeatedly photograph it multiple times and take the average as actual image to supress the noise on $y_0$.

We can easily extend the pipeline to handle color images where the transformation matrix $A$ is $\Re^{3M\times 3N}$ instead of $\Re^{M\times N}$. For calibration, just use enumerate $e_i$ three times in red, blue and green. Reconstruction is basically the same.

\section{Simulated Analysis of SparkleVision}
In this section we conduct synthetic experiments to systematically analyze how noise affects the performance of the proposed pipeline. In addition, we study how the size of mirror facet and the spatial resolution of the sparkling surface influence the resolution of the lighting that the system can recover. These experiments improve our understanding on the limits of the geometric setup, provide guidance to tune the setup, and help us interpret the results.

\subsection{Setup of the Simulated Experiment}
\begin{figure}[!tb]
\centering{
    \includegraphics[width = 0.9 \linewidth]{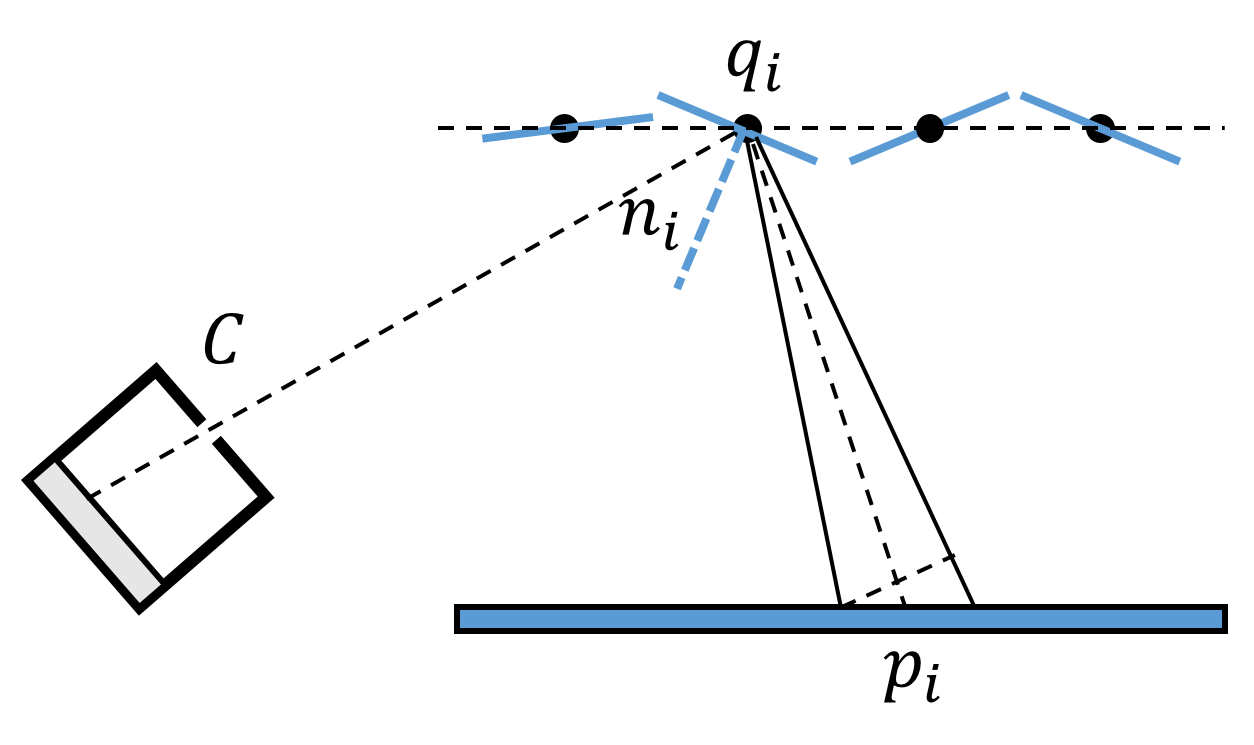}
}
\caption{Configuration of the synthetic simulation. The pixel $p_i$ with certain width is reflected by the facet $q_i$ to the camera.}
\label{fig:res_config}
\end{figure}

The setup of the simulated experiment is shown in Figure \ref{fig:res_config}, where a planar screen is reflected by a sparkling surface to the camera. The resolution of the screen is the resolution of the lightmap. We model the sparkling plane as a rectangle with fixed size. We divide the plane into blocks and each block is fully covered by a mirror facing certain orientation. The resolution of the sparkling plane is the just the number of blocks in that rectangular space. For simplicity, we do not consider interreflection or occlusion between the mirrors facets.

For each mirror facet, we assume that its orientation follows a distribution. Let $\theta\in [0, \pi/2]$ be the slant of the orientation and $\phi\in [0, 2\pi)$ be the its tilt. We model the tilt $\phi$ as uniformly distributed in $[0, 2\pi)$. In practice the mirror is centered around $0$. Therefore we model it as positive half of the Gaussian distribution with standard deviation $\sigma_\theta$. Specifically, we have
\begin{equation}
\mathrm{P}_{\theta}(\theta_0) = \frac{2}{\sqrt{2\pi}\sigma_\theta}\exp\left(-\frac{\theta_0^2}{2\sigma_\theta^2}\right), \theta_0\ge 0
\end{equation}
Note that the mean of $\theta$ is actually not $0$ and hence the actual standard deviation is not $\sigma_\theta$.

We assume that the orientation of each facet does not depend on the other so we can independently sample its value and create a random reflector. We use the classic ray-tracing algorithm to render the light reflected by the specular surface into the camera. We test the pipeline in this synthetic setup and in the ideal noise free case the recovery is perfect.

%

\subsection{Sensitivity to Noise}
\begin{figure}[!tb]
\centering{
    \includegraphics[width = \linewidth]{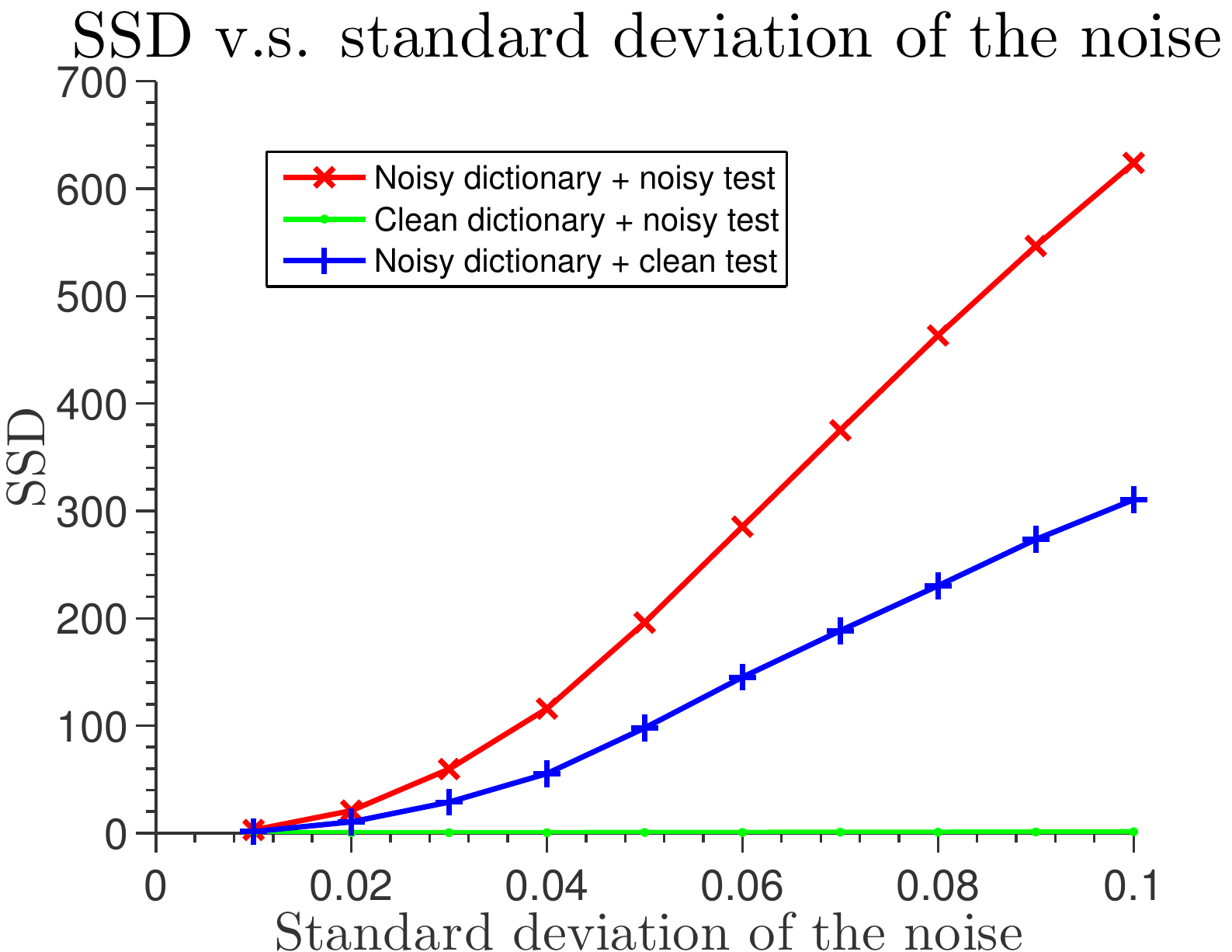}
}
\caption{How noise in the calibration and the reconstruction stage affects the recovery accuracy. Our pipeline is stable to the noise in the reconstruction stage, but not stable to the noise in the training stage.}
\label{fig:sim_noise_cmp}
\end{figure}

For simplicity, we model the image noise as i.i.d. white noise. We perform three control experiments to tease apart the effects of noise during calibration versus during test time reconstruction. In the first test, we add noise to both calibration and test images. In the second test, we only add noise to the training dictionary while in the third we only add noise to the test images. Suppose we test our pipeline on $N$ test lightmaps $I_i, 1\le i\le N$ and get the recover$\widehat{I}_i$. We measure the error of the recovery by the average sum of squared difference (SSD) between $I_i$ and $\widehat{I}_i$. Varying the noise standard deviation from $0.01$ to $0.1$, we get three curves for the tests shown in Figure \ref{fig:sim_noise_cmp}.

The result indicates that our system is much more robust to the noise in the test image than the noise in the images for calibration. In fact, when the standard deviation of the noise is $0.1$ the recovery is still great with the clean calibration images. In addition, when the noise std is as low as $0.01$, the SSD with pure the training noise is $1.85$ while the SSD with pure testing noise is just $0.14$. Therefore this comparison validates the need to use an over-complete basis in our proposed pipeline to reduce the noise in the training stage.


\subsection{Impact of Spatial Resolution}
The spatial resolution of the random reflector determines the resolution of the light map that we can recover. Keep in mind that each micro facet in our model is a mirror and our system relies on the light from the screen reflected by the facet to the camera. If some part of the screen is never reflected to the camera, there is no hope to recover from what that part of the screen is showing from the photograph taken by the camera. Since the facets are randomly oriented, this undesirable situation may well happen.

\begin{figure}[!tb]
\centering{
    \subfigure[Image shown on the screen]{
        \label{fig:spatial_orig}
        \includegraphics[height = 0.4\linewidth]{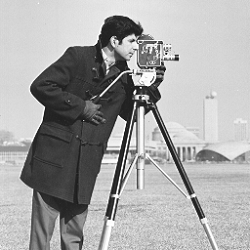}
    }
    \subfigure[Photography of a specular facet]{
        \label{fig:spatial_photo}
        \includegraphics[height = 0.4\linewidth]{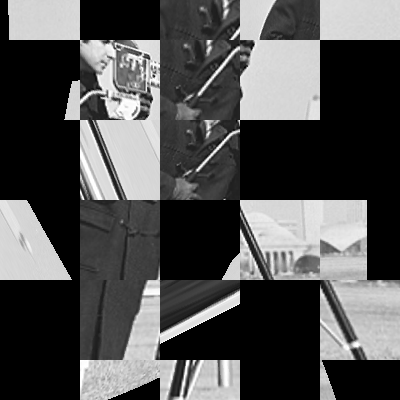}
    }
}
\caption{Photography of a specular reflector with low spatial resolution.}
\label{fig:spatial_res_teaser}
\end{figure}

Figure \ref{fig:spatial_res_teaser} demonstrates such a phenomenon. Figure \ref{fig:spatial_orig} shows a high-resolution image shown on the screen serving as the lightmap. The lightings are reflected by the micro facets to the camera sensor. However, the number of the facets is very small. As a consequence, some blocks of the photo are dark and part of the lightmap is missing, as is observed in Figure \ref{fig:spatial_photo}. Intuitively speaking, if we have more facets, the chance that part of the light map is reflected to the camera will increase, even if the size of each facet is smaller.

We develop a mathematical model to approximately compute the probability that a block of pixels on the screen will be reflected by at least one micro facet to the camera sensor. The model involves several approximations such as a small angle approximation, so the relative values in this analysis are more important than the raw values. Following the general setup in Figure \ref{fig:res_config}, we first calculate the probability that a certain pixel $p_i$ on the screen gets reflected by the micro facet $q_i$ to the camera.  Suppose the width of the pixel is $w$, then the foreshortened area of the pixel with respect to the incoming light direction $\overrightarrow{p_iq_i}$ is $w^2\cos\theta$, where $\theta$ is the angle between $\overrightarrow{q_ip_i}$ and the screen. Then the solid angle of this foreshortened area with respect to $q_i$ is $\frac{w^2\cos\theta}{\|\overrightarrow{p_iq_i}\|}$.

The normal $n$ that just reflects $\overrightarrow{p_iq_i}$ to the camera $C$ is the normalized bisector of $\overrightarrow{q_ip_i}$ and $\overrightarrow{q_iC}$. Since the incoming lights can vary in the solid angle of $\frac{w^2\cos\theta}{\|\overrightarrow{p_iq_i}\|}$, $n$ can vary in $\frac{w^2\cos\theta}{4\|\overrightarrow{p_iq_i}\|}$ and still the mirror can reflect some light from the pixel on the screen to the camera. Let $q_i \circ p_i$ be the event that the facet at $q_i$ will reflect some light emitted from $p_i$ to the camera $C$. Then its chance is the same as the probability for the orientation of the facet to be within that range, which is approximated by
\begin{small}
\begin{equation}
\mathrm{Pr}\left(q_i\circ p_i\right) = \frac{2}{\sqrt{2\pi}\sigma_\theta}\exp\left(-\frac{\theta_0^2}{2\sigma_\theta^2}\right) \frac{w^2\cos\theta}{4\|\overrightarrow{p_iq_i}\|}
\end{equation}
\end{small}

\begin{figure}[!tb]
\centering{
    \subfigure[Object $10\times 10$, Screen $5\times 5$]{
        \includegraphics[width = 0.4\linewidth]{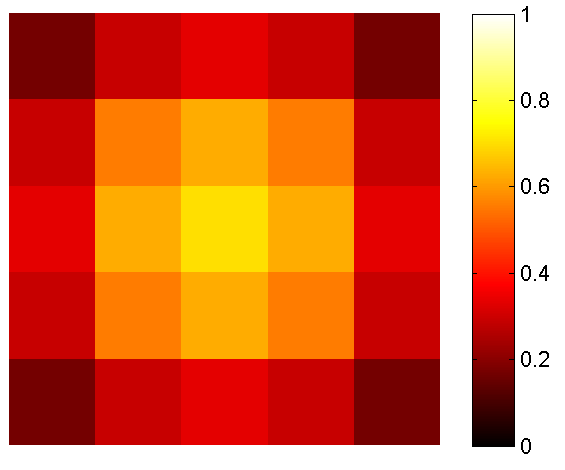}
    }
    \subfigure[Object $10\times 10$, Screen $10\times 10$]{
        \includegraphics[width = 0.4\linewidth]{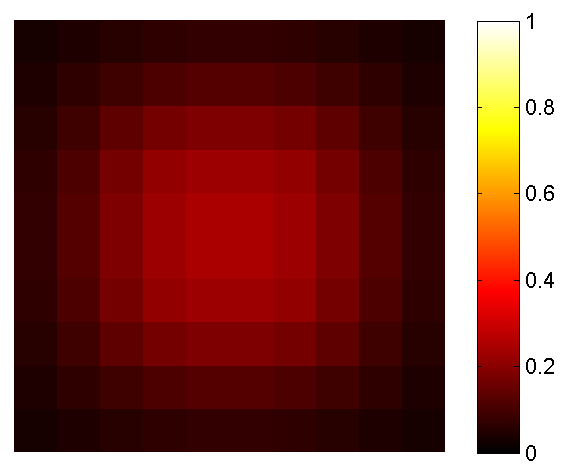}
    }\\ \subfigure[Object $20\times 20$, Screen $5\times 5$]{
        \includegraphics[width = 0.4\linewidth]{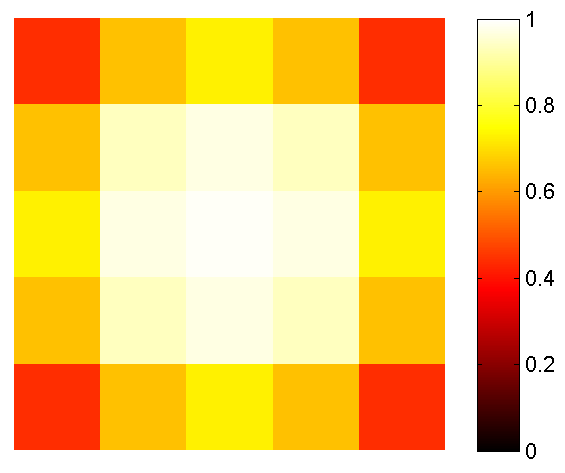}
    }
    \subfigure[Object $20\times 20$, Screen $10\times 10$]{
        \includegraphics[width = 0.4\linewidth]{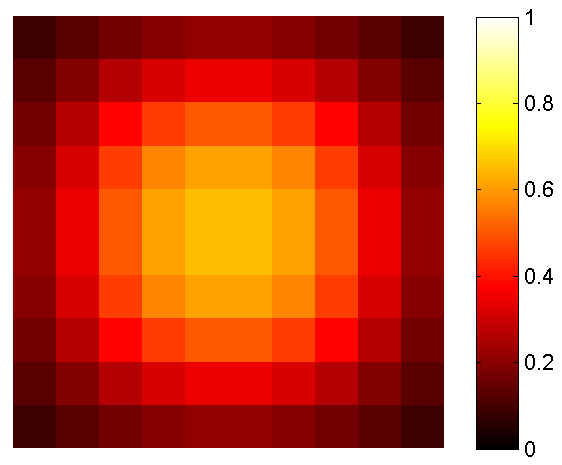}
    }
}
\caption{Probability map of light from a block pixels getting reflected by the specular surface to the screen.}
\label{fig:reflect_prob}
\end{figure}

Suppose there are $M$ micro facets in total and we compute $\mathrm{Pr}\left(q_i\circ p_i\right)$ for all $1\le i\le M$. Then we can compute the probability that the light from pixel $p_i$ is reflected by at least one micro facet to the camera as follows.
\begin{small}
\begin{equation*}
\begin{split}
\mathrm{Pr}\left(\exists j, q_j\circ p_i\right) & = 1 - \mathrm{Pr}\left(\forall j, q_j\not\circ p_i\right) = 1 - \prod_j \mathrm{Pr}\left(q_j\not\circ p_i\right)\\
& = 1 - \prod_j \left(1 - \mathrm{Pr}\left(q_j\circ p_i\right)\right)
\end{split}
\end{equation*}
\end{small}
We visualize such probability in Figure \ref{fig:reflect_prob} in four different configuration of screen and specular surface resolutions.

From the results, we can see that overall higher resolution of the specular object and lower resolution of the screen will reduce the chance that some block of pixels on the screen are not reflected to the sensor. In addition, on the same screen, the chance to avoid such bad events are different for different blocks of pixels, which is due to the different distances and relative angles between different parts of the screen and the reflector. This suggests that for a specular object there will be a limit on the resolution of the lightmap we can infer from it.

\section{Experiments}
\subsection{Experiment setup}
We place the sparkling object in front of a computer screen in a dark room and use a camera to photograph the object. The images displayed on the screen are considered as light map.  Figure \ref{fig:setup} illustrates the setup. Specifically in this experiment, we use a 24-inches ACER screen, a CANON rebel T2i camera and a set of specular objects including a hair pin, skull, and painted glitter board. The camera is placed on a heavy tripod to prevent even the slightest movement. We show that our system can reconstruct the world lighting at resolution up to $30\times 30$. At this resolution many objects, such as faces, can be easily recognized.

\subsection{Examine the assumption of the system}
\begin{figure}[!tb]
\centering{
	\subfigure[Experiment Setup]{
		\label{fig:setup}
		\includegraphics[height=0.4\linewidth]{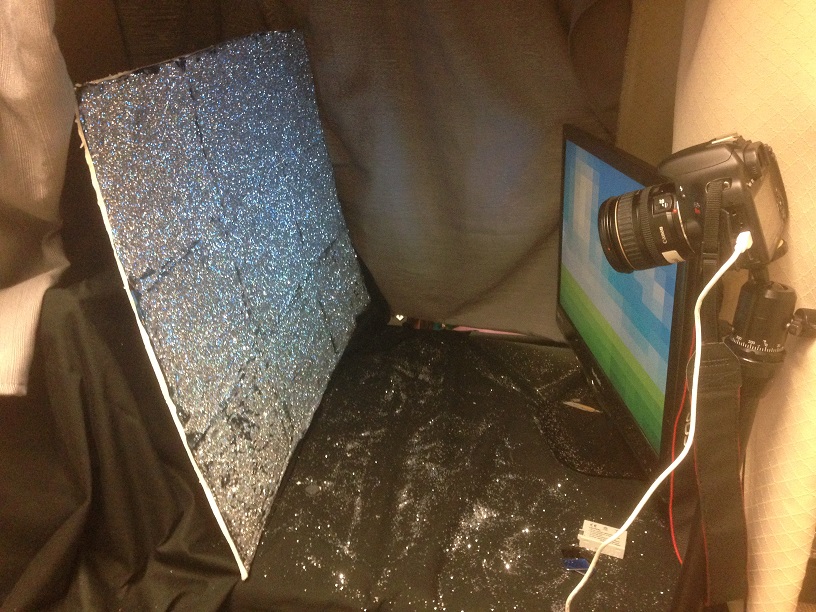}
	}
	\subfigure[Overlap graph]{
    		\label{fig:exp_overlap}
       	 	\includegraphics[height=0.4\linewidth]{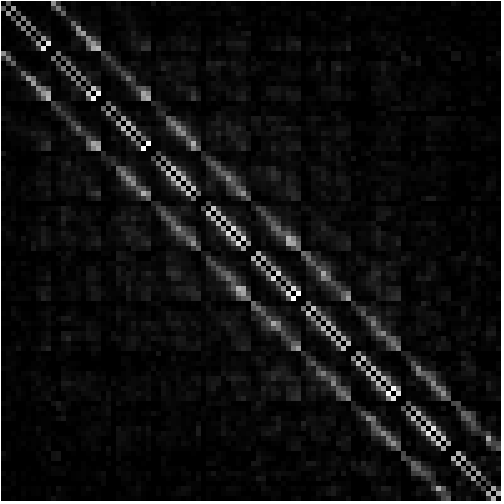}
  	  }
}
\caption{The left (a) shows the setup of the experiment in the lab. The right (b) shows the overlap graph between images from different impulses. It can be seen from the figure that only images from neighboring impulse have slight overlap.}
\end{figure}

\begin{figure}[!tb]
\centering{
    \subfigure[diffuse paper]{
    	\label{fig:diffuse_paper}
        \includegraphics[width=0.22\linewidth]{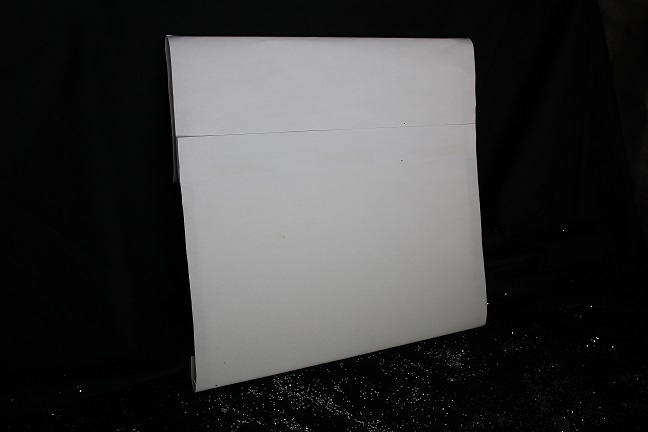}
    }
    \subfigure[glitter]{
    	\label{fig:glitter}
        \includegraphics[width=0.22\linewidth]{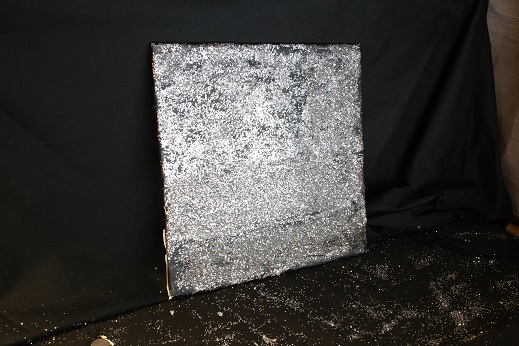}
    }
    \subfigure[hairpin]{
    	\label{fig:hairpin}
        \includegraphics[width=0.22\linewidth]{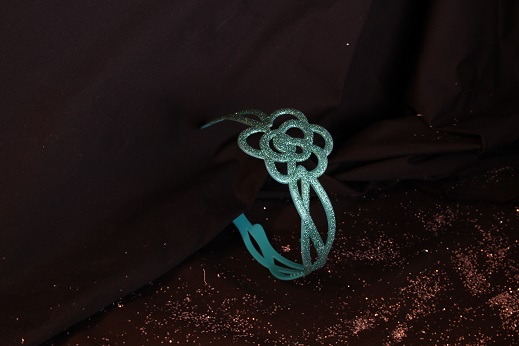}
    }
    \subfigure[skull]{
    	\label{fig:skull}
        \includegraphics[width=0.22\linewidth]{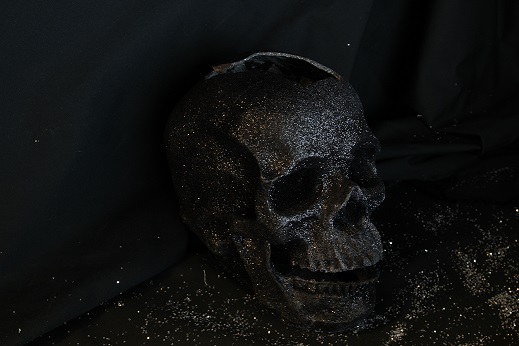}
    }\\
    \subfigure[$\kappa = 1846.31$]{
    	\label{fig:diffuse_paper_svd}
        \includegraphics[width=0.22\linewidth]{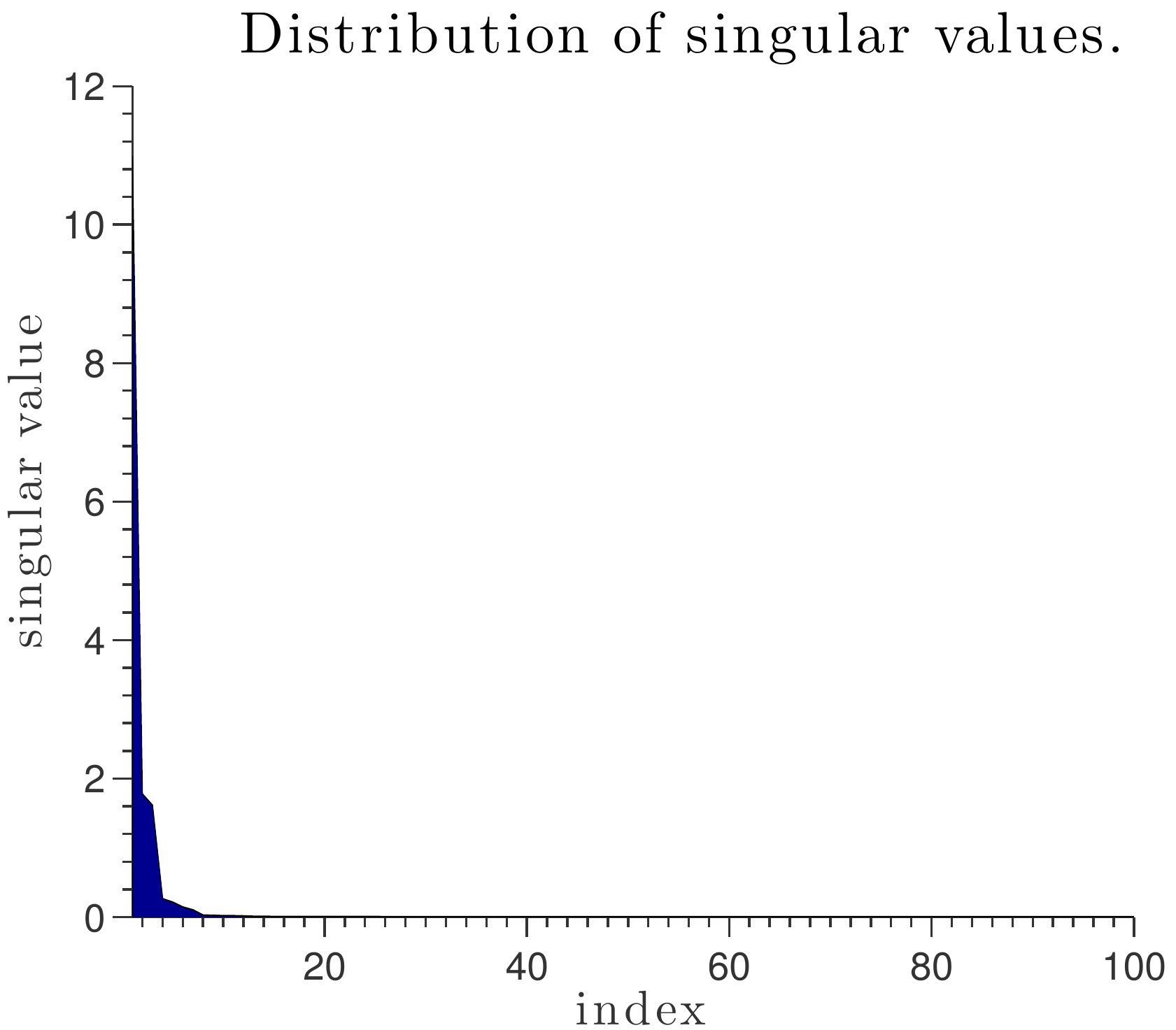}
    }
    \subfigure[$\kappa = 6.02$]{
    	\label{fig:glitter_svd}
        \includegraphics[width=0.22\linewidth]{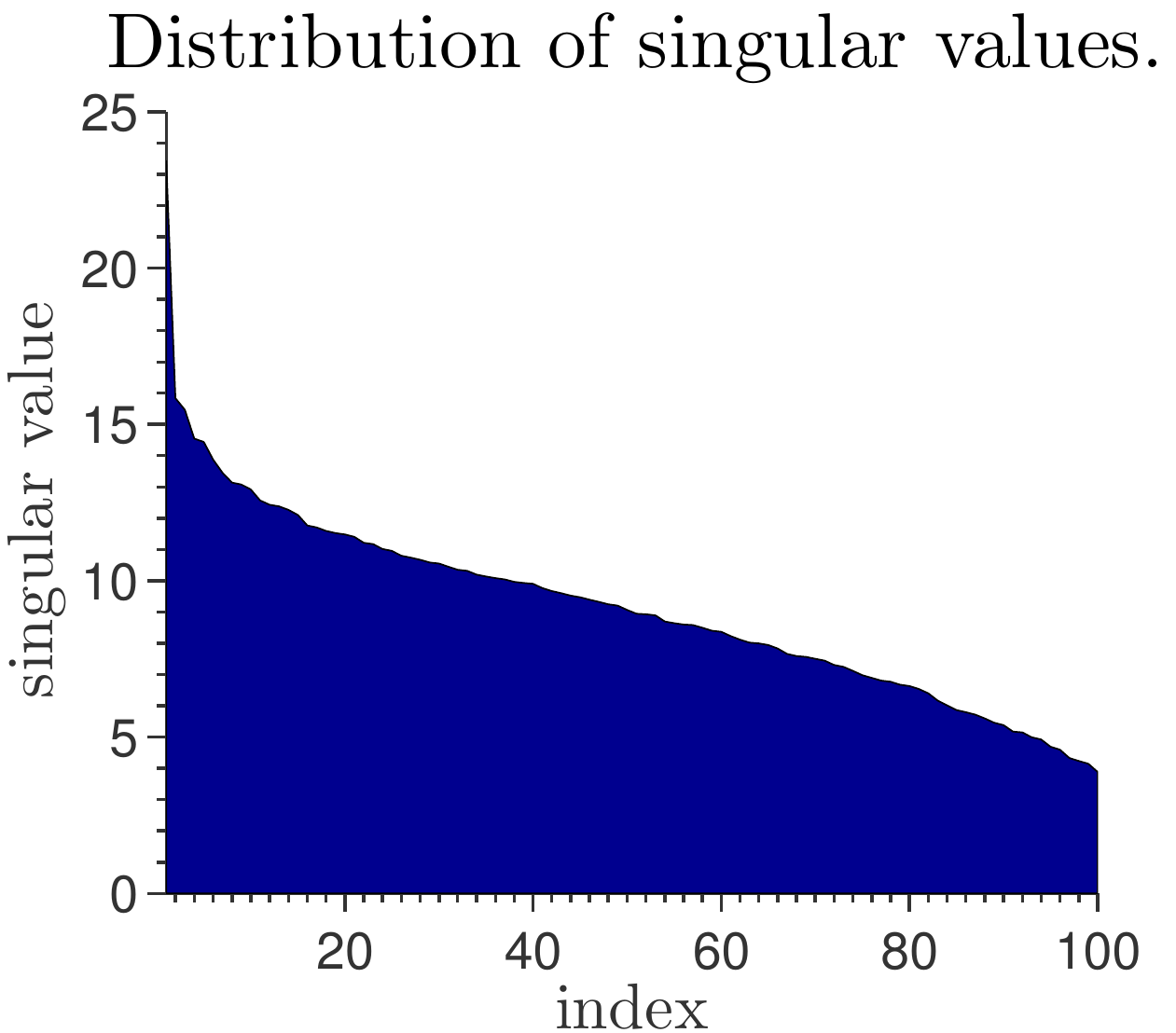}
    }
    \subfigure[$\kappa = 26.28$]{
    	\label{fig:hairpin_svd}
        \includegraphics[width=0.22\linewidth]{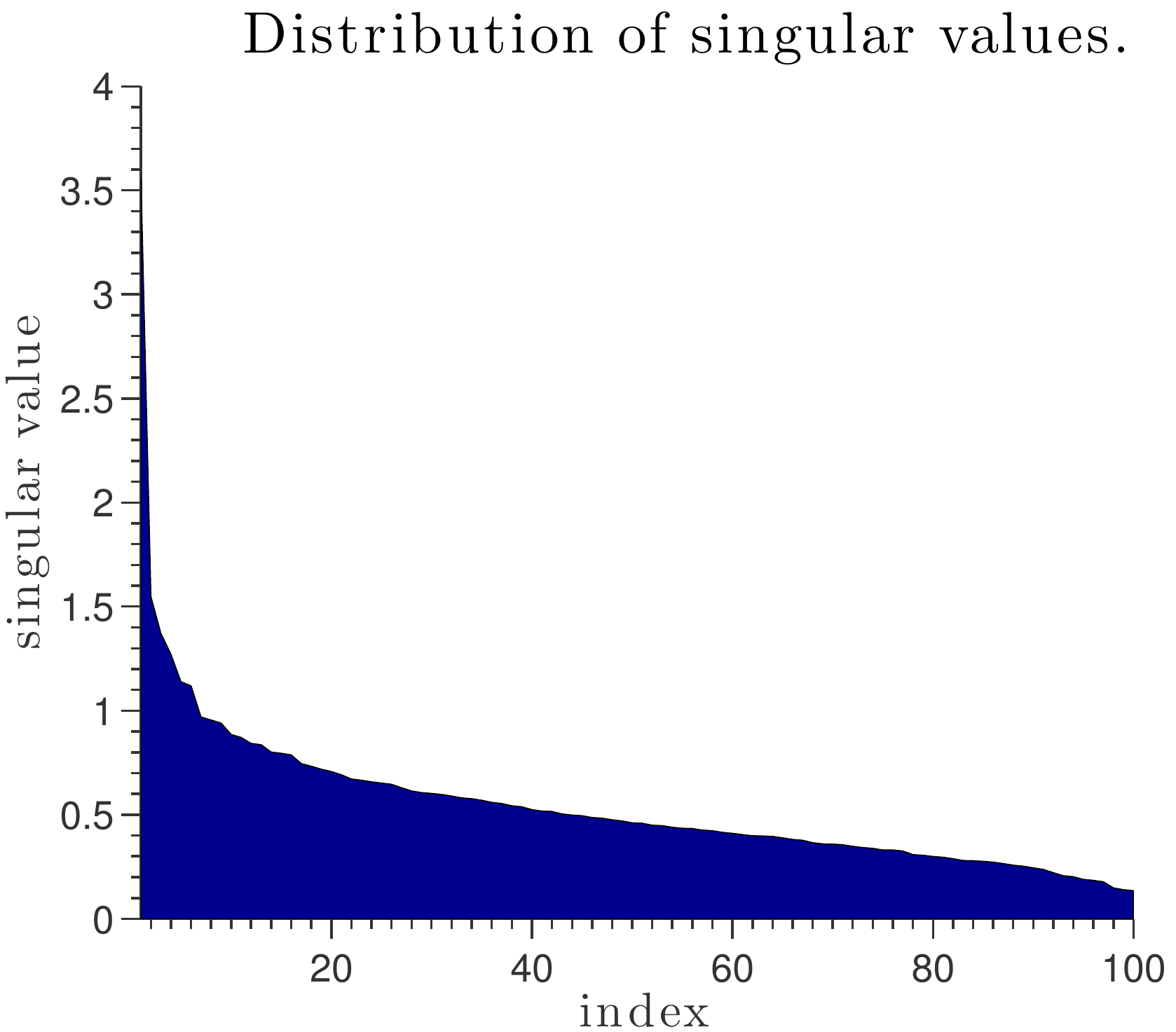}
    }
    \subfigure[$\kappa = 14.72$]{
    	\label{fig:skull_svd}
        \includegraphics[width=0.22\linewidth]{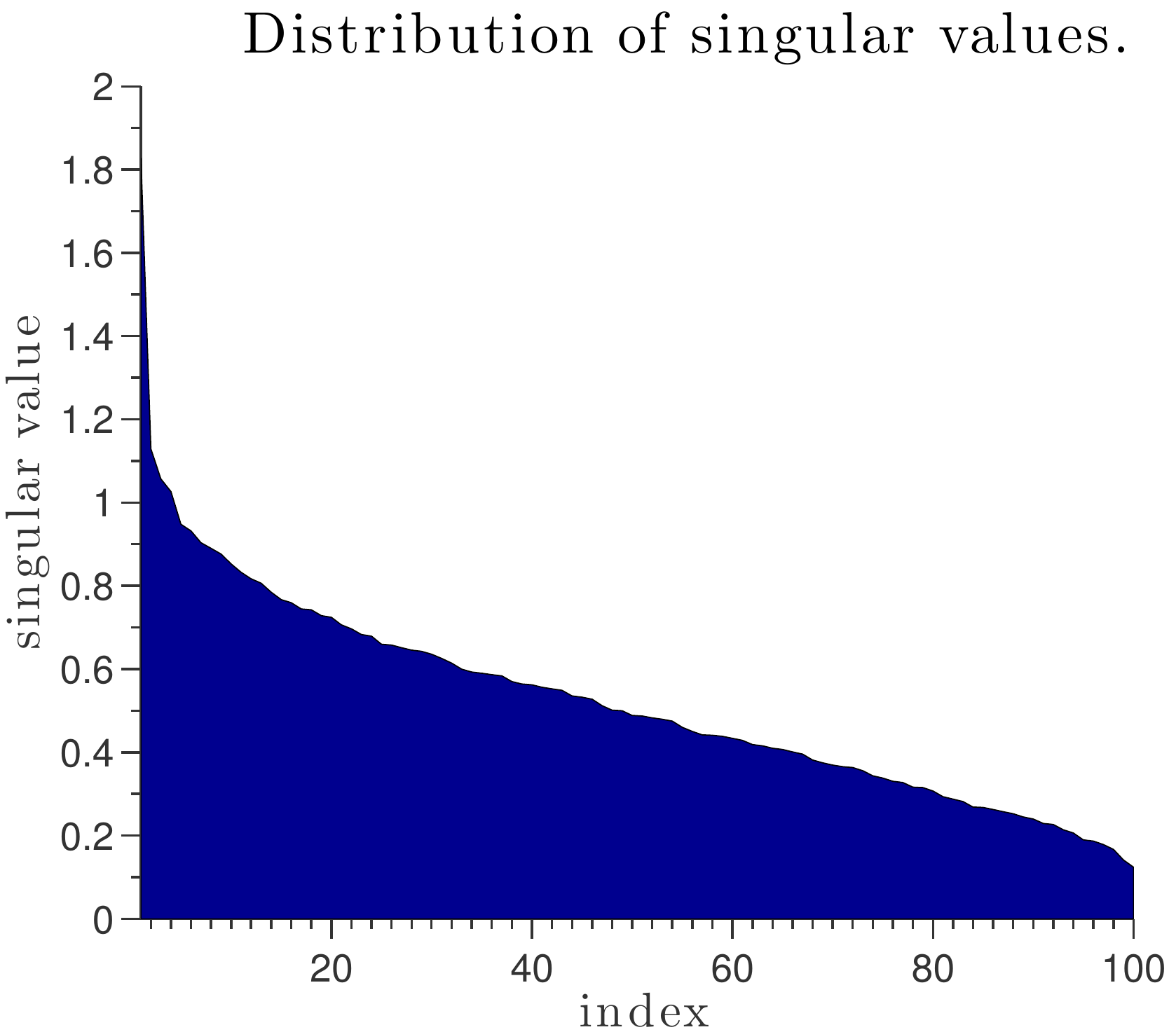}
    }
}
\caption{Singular value distribution of the transformation matrix of different objects. Also the condition number is given. Note that the sparkling reflectors create systems with much lower conditional number compared with a diffuse object.}
\label{fig:svd}
\end{figure}

\paragraph{Overlapping of bright pixels} We measure quantatively how the displacement of impulse light will change the pattern of bright spots. Let $a_i$ be the image illuminated by the impulse light and $S_i$ be the set of bright pixels in $a_i$ with intensities larger than $1/10$ of  the maximum. Then the overlap between $a_i$ and $a_j$, $i\ne j$ can be defined as
\begin{small}
\begin{equation}
\mathrm{O}(i, j) = \frac{|S_i\cap S_j|}{\min(|S_i|, |S_j|)}, i\ne j
\end{equation}
\end{small}
Here $|S|$ denotes set cardinality. At world light resolution of $10\times 10$, there are $100$ impulse basis images, and the overlap between each of them can be plotted in a $100\times 100$ image where the entry at $i$-th row and $j$-th column representing $\mathrm{O}(i, j)$, as is shown in Figure \ref{fig:exp_overlap}. As the figure suggests, most of the overlap happens between images from neighboring impulses. And the maximal overlap $\max_{i\ne j} \mathrm{O}(i, j) < 0.2$. This further validates the non-overlapping property of SparkleVision system.
\vspace{-0.2in}
\paragraph{Condition Number}
The condition number of the transformation matrix $A$, $\kappa(A)$, determines the invertibility of a transformation matrix $A$. $\kappa(A)$ is defined as the ratio between the largest and the smallest singular values of $A$. For all the optical systems shown in Figure \ref{fig:svd}, we plot all the singular values of their $A$ in descending order. From the figure, we can see that the best $\kappa(A) \approx 4$ which is good in practice.

\subsection{Real experiment results}
\begin{figure}
\centering{
\includegraphics[width = \linewidth]{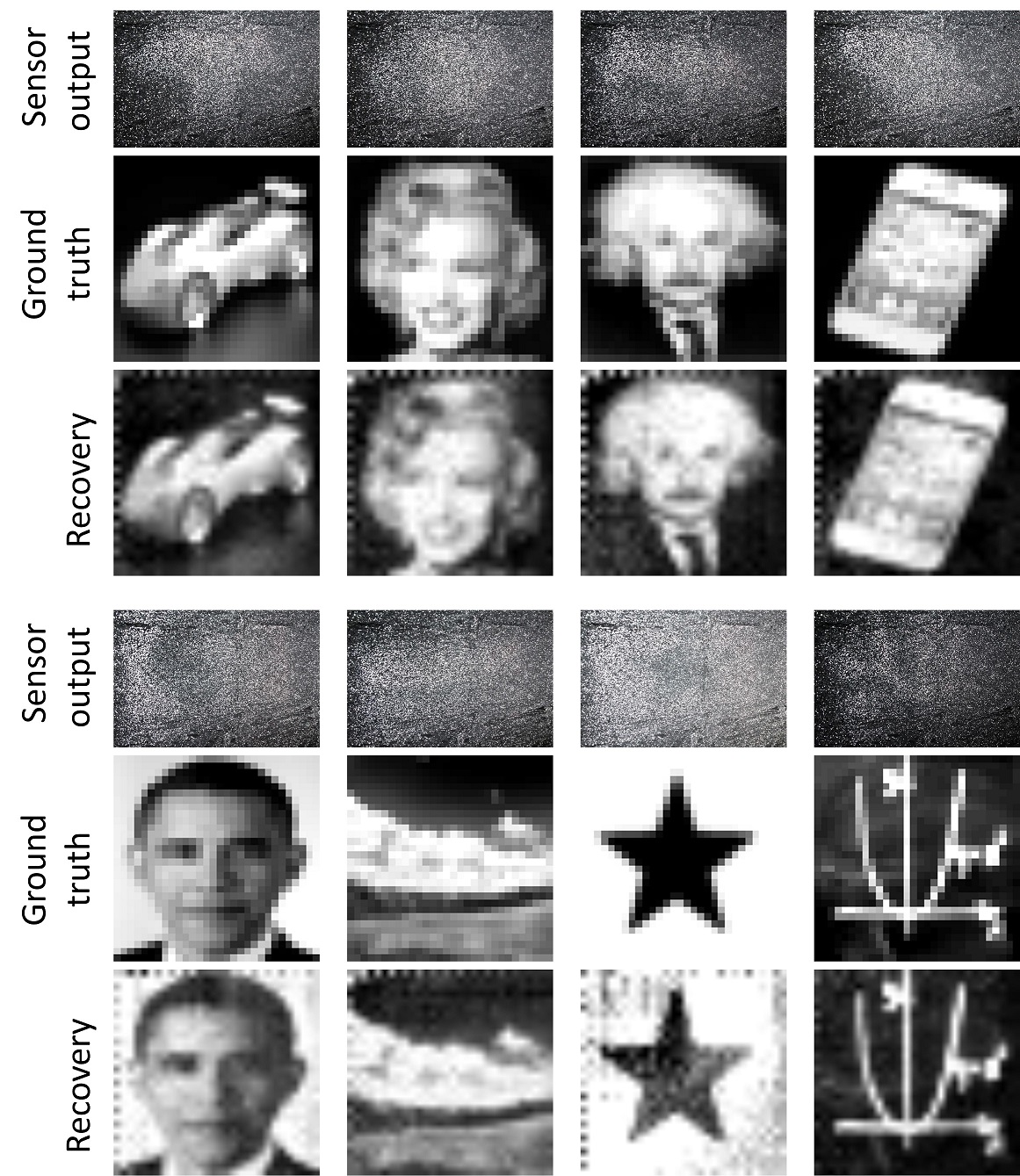}
}
\caption{SparkleVision through glitter board. Qualatitively the recovery is fairly close to the ground-truth light map and human can easily recognize the objects in the recovered image.}
\label{fig:glitter_board_result}
\end{figure}

\begin{figure}
\includegraphics[width = \linewidth]{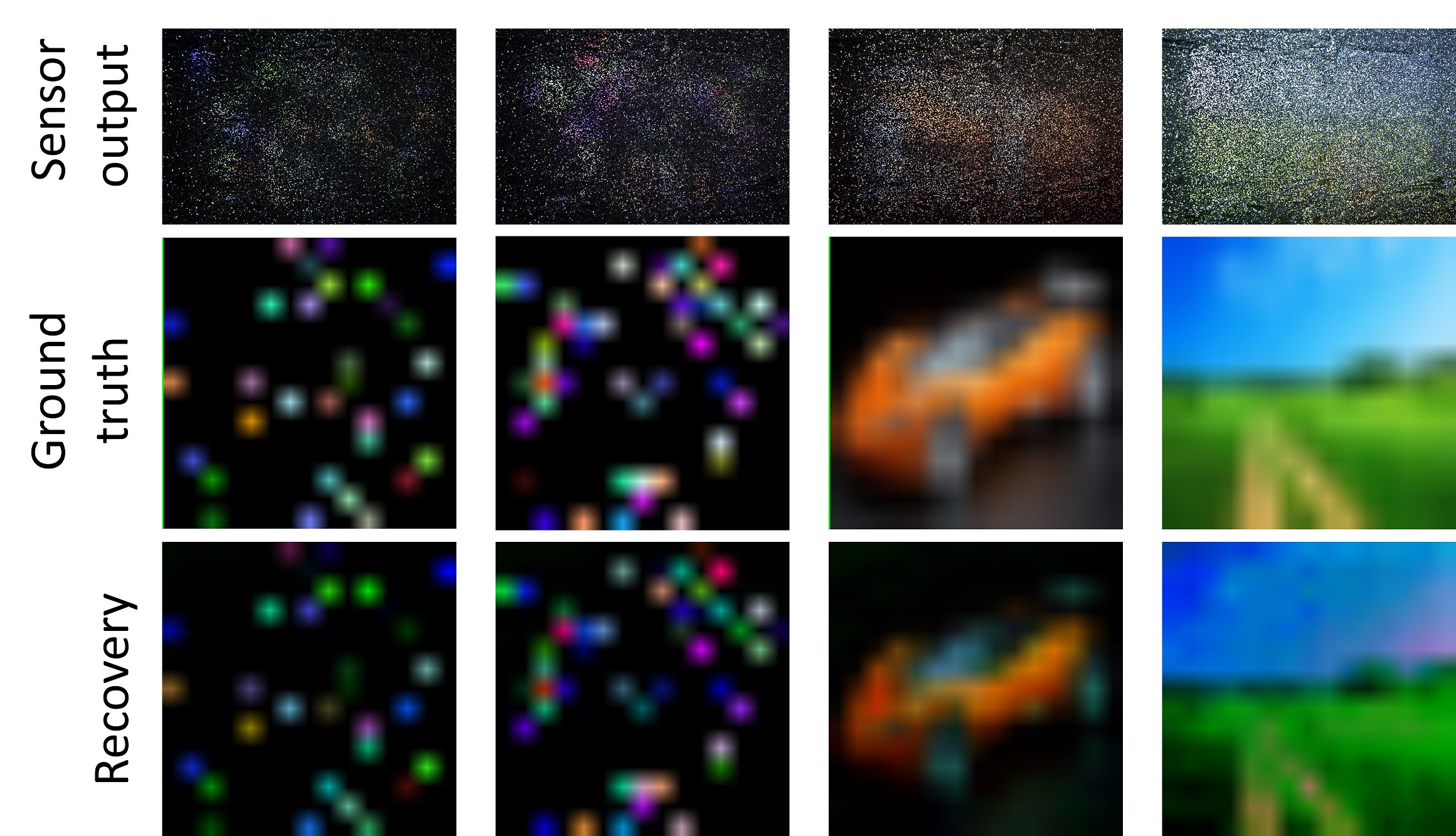}
\caption{Colored SparkleVision through glitter board. Although there is slight color distortion, the overall quality of the recovery is good.}
\label{fig:colored_glitter}
\end{figure}

Here we show results of our pipeline using the glitter as the reflector. For the gray-scale setting, we push the resolution of the screen to $30\times 30$. For the colored setting, we just present a few test at a lower resolution $15\times 15$ to demonstrate that our system generalizes. The gray scale results are shown in Figure \ref{fig:glitter_board_result} and the color ones are shown in Figure \ref{fig:colored_glitter}. They demonstrate the success of the pipeline at such resolution.

\subsection{Impact of number of basis for calibration}
\begin{figure}
\centering{
\includegraphics[width = \linewidth]{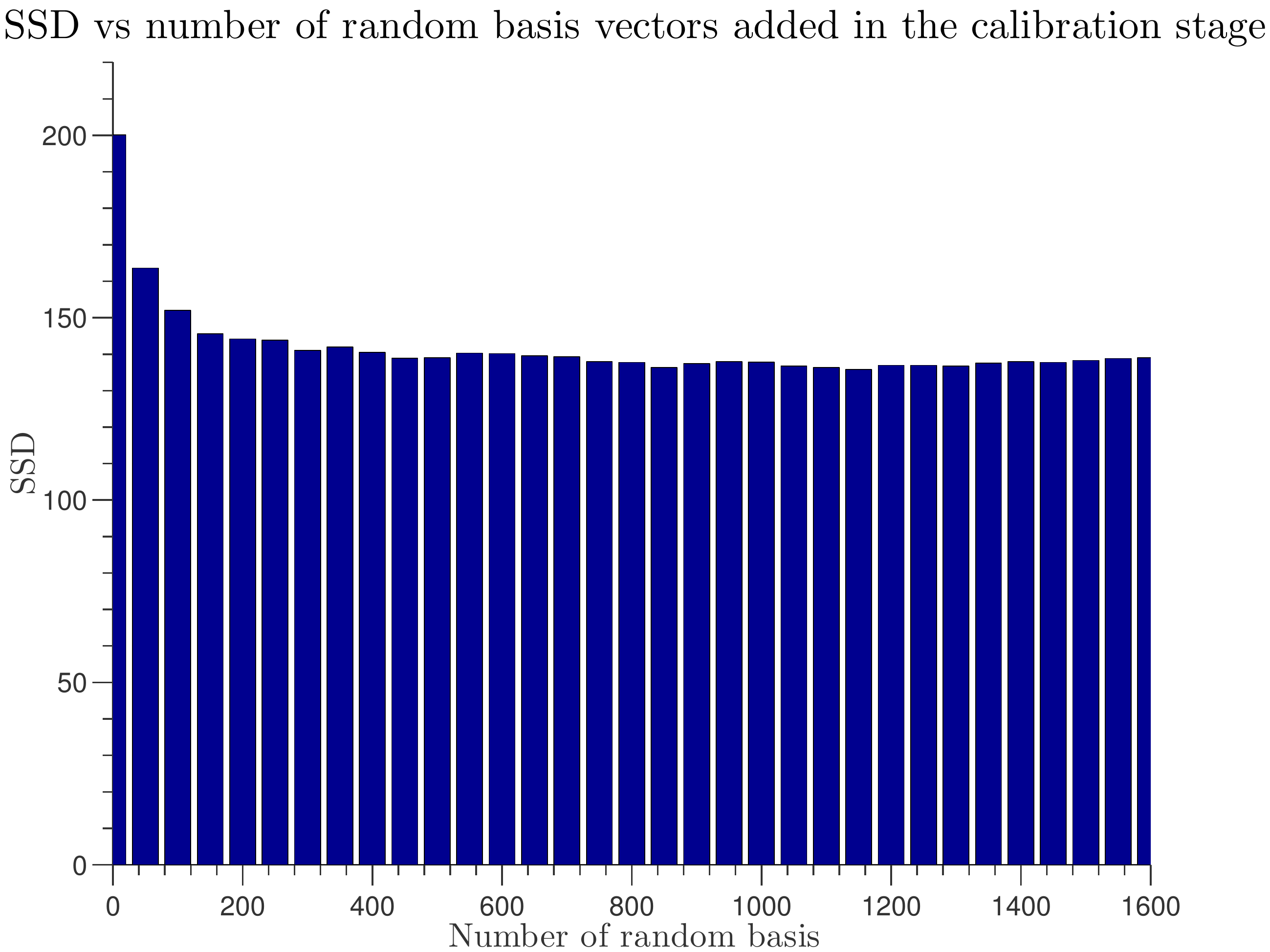}
}
\caption{Adding more random basis vectors to the calibration helps to reduce the recovery error. But this benefit saturates out.}
\label{fig:num_basis}
\end{figure}

Figure \ref{fig:num_basis} illustrates how the increasing number of random basis used in the calibration improve the recovery of the light $x$.  Note that the resolution of the light in this setup is $20\times 20$, hence the number of impulse and DCT bases are both $400$. It is worth noting that the benefit gradually saturates out so we only need to employ a limited number of random basis.

\subsection{Stability to noise}
\begin{figure}
	\subfigure[No noise]{
		\includegraphics[width=0.17\linewidth]{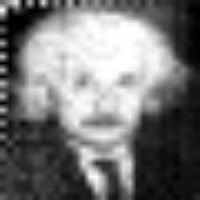}
  	}
	\subfigure[0.04]{
		\includegraphics[width=0.17\linewidth]{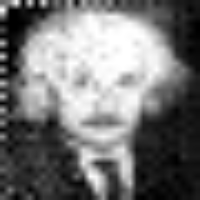}
  	}
	\subfigure[0.08]{
		\includegraphics[width=0.17\linewidth]{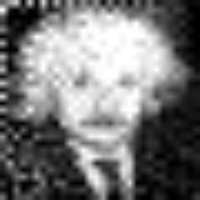}
  	}
	\subfigure[0.12]{
		\includegraphics[width=0.17\linewidth]{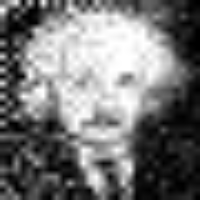}
  	}
	\subfigure[0.16 ]{
		\includegraphics[width=0.17\linewidth]{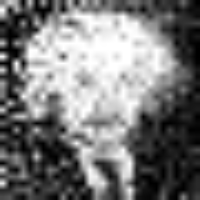}
  	}
\caption{Stability to noise on the test image: title of the subfigure represents noise level. The noise is large considering the images are in $[0, 1]$ and only a few spots are bright.}
\label{fig:noise}
\end{figure}

We perform synthetic experiments by adding noise to the real test image to understand how robust the real calibrated transformation matrix is. We measure the robustness by Root-mean-squared-error (RMSE) between the noisy recovery and the non-noisy recovery. We plot how the reconstructed lighting change as the noise level increases in Figure \ref{fig:noise}. The result validates the robustness of our system.

\subsection{Sensitivity to misalignement and potential application}
\label{sec:misalignment}
\begin{figure}
	\subfigure[1 pixel ]{
		\includegraphics[width=0.17\linewidth]{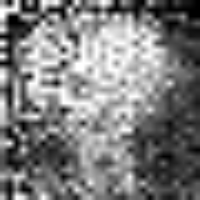}
  	}
	\subfigure[0.8 pixel]{
		\includegraphics[width=0.17\linewidth]{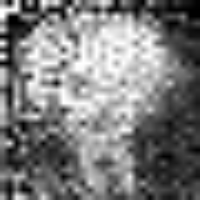}
  	}
  	\subfigure[0.4 pixel]{
		\includegraphics[width=0.17\linewidth]{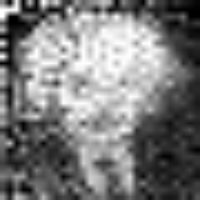}
  	}
	\subfigure[0.2 pixel]{
		\includegraphics[width=0.17\linewidth]{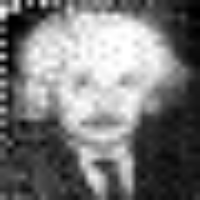}
  	}
	\subfigure[No shift]{
		\includegraphics[width=0.17\linewidth]{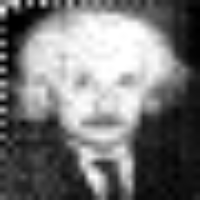}
  	}
\caption{Instability to misalignment: even if we shift the test image by one pixel horizontally, there is significant degrade in the output. We can compensate this by grid search and image prior.}
\label{fig:shift}
\end{figure}

The success of SparkleVision relies largely on the sensitivity of light pattern on a specular object to even a slight movement of the source light. However, this property simultaneously make the whole system extremely sensitive to subtle misalignment. To show this we perform synthetic experiments by shifting the test image $I$ by $\Delta x$ and examine the RMSE. Some representative results and the RMSE curve are shown in Figure \ref{fig:shift}. We could compensate for this misalignment by performing grid search over $\Delta x$ and pick out the best recovery which has minimum value of total variation $\sum_x \|\nabla I(x)\|_1$.

For the recovery of light, this phenomenon is harmful. But such sensitivity to even subpixel misalignment can enable the detection and magnification of motion of the object that is invisible to the eyes, like \cite{Wu12Eulerian}. We leave this for future work.


\section{Discussions and Conclusion}

In this paper we show that it is possible to infer an image of the world around an object that is covered in random specular facets. This class of objects actually provide rich information about the environmental map and is significantly different from the smooth objects with either Lambertian or specular surfaces, which researchers in the field of shape-from-X have worked on.

The main contributions of the paper are twofold. First, we have presented the phenomenon that specular random microfacets can encode a large amount of information about the surrounding light. This property may seem mysterious at the first sight but indeed is intuitive and simple once we understand it. We also analyze the factors that affect the optical limits of these reflectors. Second, we proposed and analyzed a physical system that can efficiently perform the calibration and inference of the surrounding light map based on these sparkling surfaces.

Currently our approach only reconstructs a single image of the scene facing the sparkling object. Such an image corresponds to a slice of the lightfield around the object. Using an identical setup, it should be possible to reconstruct other slices of the lightfield. Thus, our system could be naturally extended to work as a lightfield camera. In addition, this new reflector has the ability of reveal subtle motions of the optical setup. We leave all these exciting directions for future exploration.


{\small
\bibliographystyle{ieee}
\bibliography{sparkle}
}

\end{document}